\documentclass[10pt, journal]{article}
\usepackage{blindtext}
\usepackage{graphicx} 
\graphicspath{{/home/mike/Dropbox/Latex/Graphics/}}
\DeclareGraphicsExtensions{.png, .jpg, .pdf}
\usepackage{amsmath} 
\usepackage{amsfonts}
\usepackage{bbm}
\usepackage{subfig}
\usepackage{float}
\usepackage[utf8]{inputenc}
\usepackage{graphicx}
\usepackage{lipsum}
\usepackage[margin={2cm,2cm}]{geometry}
\usepackage{amsthm}
\theoremstyle{plain}
\newtheorem{theorem}{Theorem}
\usepackage{url}
\usepackage{tabularx}
\usepackage{multicol}
\usepackage{vwcol}
\usepackage{wrapfig}
\usepackage{authblk}

\begin{document}

	
%
\title{Monge's Optimal Transport Distance for Image Classification}
\date{}
\author[]{Michael Snow}
\author[]{Jan Van lent}
\affil[]{\small\textit{Department of Engineering Design and Mathematics, Centre for Machine Vision,} \\ \small\textit{University of the West of England, Bristol, BS16 1QY}}
\affil[]{\textit{Michael1.Snow@outlook.com}, \textit{Jan.Vanlent@uwe.ac.uk}}

\maketitle

\begin{multicols}{2}

\begin{abstract}
	
\textit{This paper focuses on a similarity measure, known as the Wasserstein distance, with which to compare images. The Wasserstein distance results from a partial differential equation (PDE) formulation of Monge's optimal transport problem. We present an efficient numerical solution method for solving Monge's problem. To demonstrate the measure's discriminatory power when comparing images, we use a $1$-Nearest Neighbour ($1$-NN) machine learning algorithm to illustrate the measure's potential benefits over other more traditional distance metrics and also the Tangent Space distance, designed to perform excellently on the well-known MNIST dataset. To our knowledge, the PDE formulation of the Wasserstein metric has not been presented for dealing with image comparison, nor has the Wasserstein distance been used within the $1$-nearest neighbour architecture.}
\end{abstract}


\section{Introduction}

The problem of optimal mass transport was first introduced in a seminal paper by Gaspard Monge, \textit{M$\acute{\textit{e}}$moire sur la th$\acute{\textit{e}}$orie des d$\acute{\textit{e}}$blais et des remblais}, in 1781 \cite{GM01}. In recent years the theme of optimal transport has attracted researchers from a number of areas where its link to their subject has been realised; including economics, fluid mechanics and image processing. Despite its prominence in many mathematical applications, optimal transport seems to be little known in the image processing community where it could potentially be of significant benefit in image comparison techniques. Monge's optimal transport problem and the associated Wasserstein distance metric, defined in Section II, give a way to measure the similarity between probability distributions. If we consider two normalised images to be probability distributions, then we can think of the Wasserstein distance as a measure of the most efficient way to transport one image onto the other, based on some measure of cost from transporting a pixel from one image to another. The intuition for using such a metric is natural; if we consider two images to be visually similar, then it is sensible to think that the distribution of the pixels will also be similar. In certain situations, computing the optimal transport distance between the distributions can be more beneficial than using other traditional distance metrics.

\begin{figure}[H]

	\includegraphics[width=0.45\textwidth]{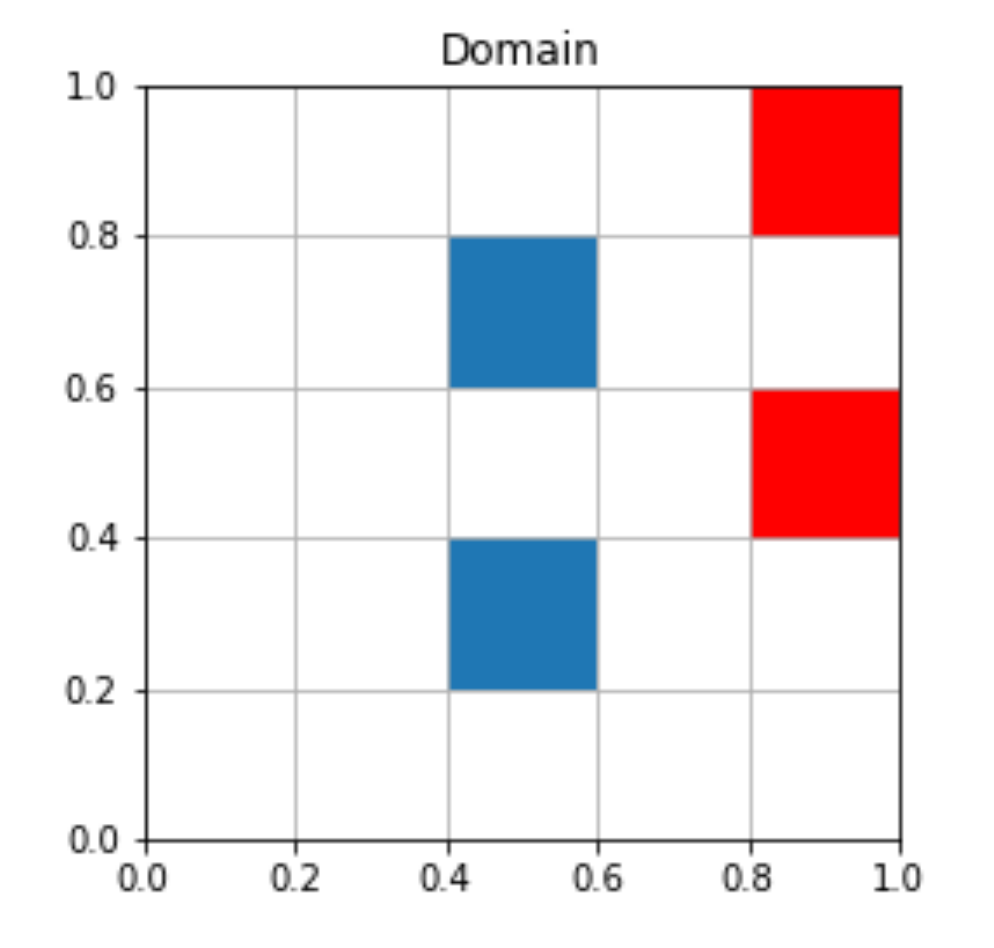}
	\caption{Image Comparison Example.}
	\label{binary}

\end{figure}

To introduce and motivate the idea of optimal transport, we shall provide a simple example on which we compare the commonly used Euclidean distance to a discrete formulation of the optimal transport problem; sometimes called the Kantorovich distance. Consider two images: one an inverted binary $5 \times 5$ image, where black has a pixel value $1$ and white has a pixel value $0$. The second image is a translation of the first; this is presented in Figure \ref{binary}, where for clarity we've overlapped both images, where the black pixels of image one and image two are coloured blue and red respectively. For the purposes of optimal transport, we shall say these images both lie in the domain of the unit square. Visually these images are similar to us; the are simply a translation. One commonly used way to measure similarity between images is the Euclidean distance. If we consider that each image can be represented as a vector, where the columns of the images are concatenated from left to right, then an image can be represented by $\mathbf{v} \in \mathbb{R}^{n}$, such that $\mathbf{v} = \left( v_1, v_2, \, . \, . \, , v_{n-1}, v_{n} \right)^T $ and each component of $\mathbf{v}$ corresponds to a pixel intensity value. The Euclidean distance between two vectors $\mathbf{x}$ and $\mathbf{y}$ is defined to be 

\begin{align}
d(\mathbf{x}, \mathbf{y}) = \sqrt{(x_1 - y_1)^2 +  \cdot \cdot  (x_n - y_n)^2}.
\end{align}

This measures the pixel-wise distance between the images; if the pixel intensities between the images are similar then the Euclidean distance can give a good metric. However, in this case, the metric would give a poor representation of how visually similar the images actually are; the areas of interest are disjoint between the images, so despite being visually similar this gives a distance of $2$ which is the worst possible result as no black pixels overlap. 

We can instead think of this as a transport problem. To think of this as a transport problem it is conceptually easier to think of the blue pixels in the first image as factories and the red pixels in the second image as retailers. The pixel intensities $0$ and $1$ represent the stock of the factory in the first image and the demand from the retailers in the second image. We shall assume that the supply is equal to the demand and that to transport the goods from a factory to a retailer has a cost attached which is related to the distance between them. The optimal transport problem is then, "What is the optimal way to transport the goods from the factories to the retailers such that the total cost is minimised based on the distance?" 

This discrete problem, known as the discrete Monge-Kantorovich optimal transport problem, can be formulated as a finite-dimensional linear programming problem \cite{LE01}. We shall denote the images in vector form so that the images to be compared are denoted by the vectors $\mu_i^+$ and $\mu_j^-$ respectively, for $i = 1, . .. \, , n$ and $j = 1, ... \, , n$. For the problem to be valid, the total pixel sum of the images have to be equal so that 

\begin{align}
\sum_{i=1}^{n} \mu_i^+ = \sum_{j=1}^{n} \mu_j^-.
\end{align}

Clearly this is the case as observed in Figure \ref{binary}; both images sum to the value of 2. We must also construct a cost vector $c_{ij}$. The vector $c_{ij}$ measures a chosen distance between the pixels $\mu_i^+$ and $\mu_j^-$ at the points $i$ and $j$ according to the original square geometry. This is to measure the cost of transporting a pixel from one image to another. In this example, we shall assume that the horizontal and vertical distance between neighbouring pixels is a single unit and we choose our distance function to be the squared distance $c_{ij} = \lvert i - j \rvert^2$, where $i$ and $j$ in this case denote the pixel coordinates. The problem then becomes a minimisation: we wish to find some transport plan $\mu_{ij} \geq 0$ so that we minimise

\begin{align}
\sum_{i=1}^n \sum_{j=1}^n c_{ij} \mu_{ij}, 	
\end{align}

subject to the constraints

\begin{align}
\sum_{j=1}^n \mu_{ij} = \mu_i^+ \qquad \text{and} \qquad \sum_{i=1}^n \mu_{ij} = \mu_j^-,  
\end{align}

for $i = 1, ... \, n$, $j = 1, ... \, n$ and $\mu$ strictly non-negative. This can be written as a standard primal linear programming (LP) problem \cite{LE01}. If $x = \left(\mu_{11}, \mu_{12}, ... \, , \mu_{nn}\right)^T$, then the LP formulation is

\begin{align*}
&\textbf{Minimise} \,\, c \cdot x, \,\, \text{subject to the constraints} \\
&Ax = b, \,\, x \geq 0, 
\end{align*}

where $b = \left(\mu_1^+, ... \, , \mu_n^+, \mu_1^-, ... \, , \mu_n^-\right)^T$  and $A$ is the constraint matrix of the form 

\begin{align}
A = \begin{bmatrix}  \mathbbm{1} & 0 & \dots & 0 \\
                 0 & \mathbbm{1} & \dots & 0 \\
                 \vdots & \vdots & \dots & \vdots \\
                 0 & 0 & \dots & \mathbbm{1} \\
                 e_1 & e_1 & \dots & e_1 \\
                 e_2 & e_2 & \dots & e_2 \\
                 \vdots & \vdots & \dots & \vdots \\
                 e_n & e_n & \dots & e_n \\
                 \end{bmatrix}.
\end{align}

\begin{figure}[H]
	\includegraphics[width=0.45\textwidth]{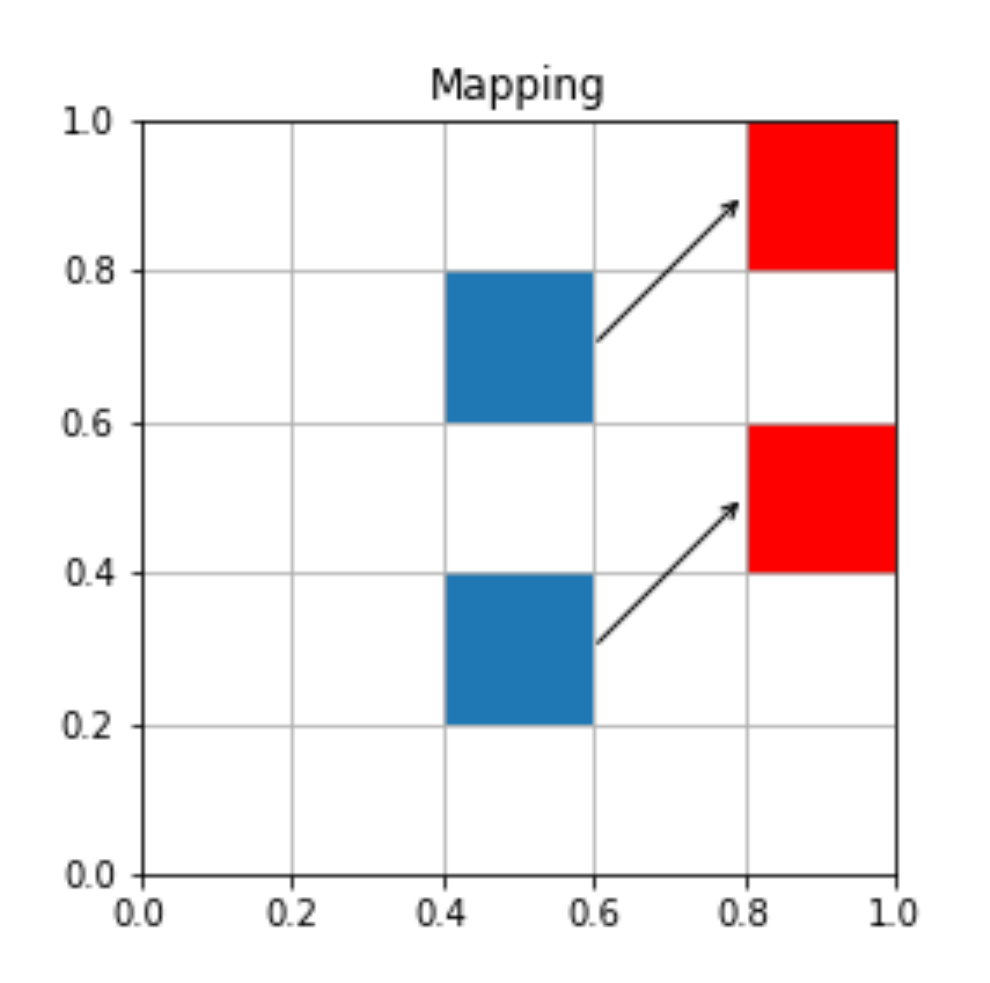}
	\caption{Example of Optimal Transport.}
	\label{OTexample}
\end{figure}

Here, $\mathbbm{1} \in \mathbb{R}^n$ is a row vector of ones and $e_i \in \mathbb{R}^n$ is a row vector which contains a $1$ at the $i$th position, else contains a zero. This is a standard LP problem and can be solved using standard methods, with for example an interior point method. In this case the solution is trivial enough to do by hand. Clearly the most efficient way to transport the goods is by the natural translation. This is illustrated in Figure \ref{OTexample}.

From this optimal mapping, a natural distance between the images can be used. To measure the similarity we consider the total cost of transporting the goods under the optimal transport plan, $\mu$. So we define the Kantorovich distance to be 

\begin{align}
K = \sum_{i=1}^n\sum_{j=1}^n \mu_{ij} c_{ij}.
\end{align} 

In our example, each of the two pixels have a mass of $1$ and the translation distance from each factory to retailer is $\sqrt(0.4^2 + 0.2^2) \approx 0.45$, so the Kantorovich distance is simply twice this and the total cost is  $\approx 0.9$. It should be noted that this scalar is not comparable with the Euclidean distance; each distance is relative to its own measure. 

In this case it can be argued that the Kantorovich distance is a more natural way to think of the distance between two images as it can be thought of as the optimal way to map one topology onto another. Indeed there are potentially many situations in the fields of image processing, computer vision and machine learning where the optimal transport distance could be a useful tool; and this provides the motivation for this paper. It should be noted that there are indeed image metrics that do deal with the issues outlined in this simple problem; for example the tangent space metric derived by Simard et al. \cite{PS01}. In particular applications, there is certainly an argument for using an optimal transport distance. For a more complete discussion on the topic of optimal transport and recent developments, we refer the reader to Evans \cite{LE01}, Villani \cite{CV01,CV02} and McCann \cite{RM01}.

\section{Related Work}

We have found there to be a limited amount of literature on using the optimal transport distance as an image comparison measure outside of image registration; in particular using Monge's PDE formulation. There is more literature available on the rather more well-known Earth Mover's distance; see for example Ling \cite{HL01} and Rubner \cite{YR01}. This is the one-dimensional version of the Wasserstein distance and typically used to compare histograms. The literature available for the two-dimensional problem focuses on the discrete Kantorovich distance, outlined earlier in this section, rather than the PDE formulation we have used in this paper. The PDE formulation is seen more commonly in other disciplines, such as for example  meteorology \cite{EW01}. Possibly the first use of the Wasserstein distance in two dimensions is by Werman et al. \cite{MW01, MW02}. In this case Werman used a discrete `match distance' which is the discrete Kantorovich distance and demonstrated a case where images of equal pixel sum could be compared. Werman noted that the distance has been shown to have many theoretical advantages but a significant drawback is the computational cost. For instance naively comparing two $n \times n$ images would be of complexity $O(N^3)$ according to Werman, where $N$ is the number of pixels. In the paper by Kaijser \cite{TK01}, Kaijser managed to reduce the complexity of the discrete Wasserstein algorithm from $O(N^3)$ to roughly $O(N^2)$, where $N$ denotes the number of pixels, by considering \textit{admissible arcs} in which to transport from one image to another. Kaijser's paper focuses on constructing a more efficient algorithm rather than the relative benefits of using the Wasserstein distance, but he does illustrate the distance being used to compare images. In a similar vein, Alexopoulos \cite{CA01} claims an improvement upon Kaijser's algorithm and also presents some example image comparisons. There have been some notable contributions to the field of optimal transport and image processing by Delon et al. \cite{JD01, JD02}. Delon has implemented some novel approaches to transporting the colour maps from one image onto another. Delon has also worked in developing fast solvers to the transport on different domains, with some applications to images.

In our paper we will be using a PDE formulation of the Wasserstein distance, which can be found from solving a Monge-Amp\`ere equation resulting from the formulation of Monge's optimal transport problem with a quadratic cost function as set out in Section III. There are some significant benefits to using the PDE formulation of the Wasserstein distance over the discrete version. As seen in the literature, the discrete Wasserstein distance posed as a linear programming transport problem has a naive cost of $O(N^3)$, improved by Kaijser to $O(N^2)$. When comparing two images, it can quickly become impractical to compute this distance in finite time. With the PDE formulation, the numerical solution method we have used results in iteratively solving a 2nd order linear elliptical PDE with variable coefficients. In each iteration this involves solving a large sparse system of linear equations. By exploiting the sparsity of the problem, we would expect our algorithm to have a complexity between $O(N)$ and $O(N^2)$; in Section IV on an example we tried, the computational cost is approximately $O(N^{1.7})$. The PDE formulation could also potentially be more accurate if some sensible assumptions are made as to how to represent a discrete image as a continuous function. 

\subsection*{Contributions and Outline}

The main contributions of this paper are as follows. The paper sets out a numerical solution method to solving Monge's PDE formulation of the optimal transport problem and from the numerical solution, we have a numerical approximation to the Wasserstein distance as defined in Section III. The paper also illustrates this optimal transport distance in a machine learning context; specifically the $1$-NN framework to show the discriminative power of the distance. Finally we compare this optimal transport distance to the Euclidean distance, the Pearson's correlation coefficient and finally the Tangent Space distance on the well-known MNIST dataset. We show that in this case, the optimal transport distance performs at an impressive level. 
  
We set this out as follows. In the following section, we present the theoretical framework of the Monge's optimal transport problem in Euclidean geometry, based on a quadratic cost function. In Section IV we present a numerical solution method to the Monge-Ampere equation based on a standard finite difference scheme. In Section V we introduce the $k$-nearest neighbour ($k$-NN) algorithm using the Wasserstein distance as the distance metric. 
In Sections VI we use the well-known MNIST optical character recognition data set to illustrate the potential benefits of using the optimal transport distance over two traditionally used metrics and the state-of-the-art Tangent Space distance. Finally in Sections VIII and IX, we summarise our results, future work; then give our acknowledgements. 

\section{Monge's Optimal Transport Problem in Euclidean Geometry}

Gaspard Monge was the founding father of the study of optimal transport. Monge's problem is as follows: consider a finite pile of rubble (mass) and a hole of the exact same volume. Assume that to transport the dirt to the hole requires some measure of effort, then Monge's problem is to find an optimal mapping to transport the mass into the hole such that the effort is minimised \cite{GM01}. 

\begin{figure}[H]
    \includegraphics[width=0.45\textwidth, height = 0.15\textheight]{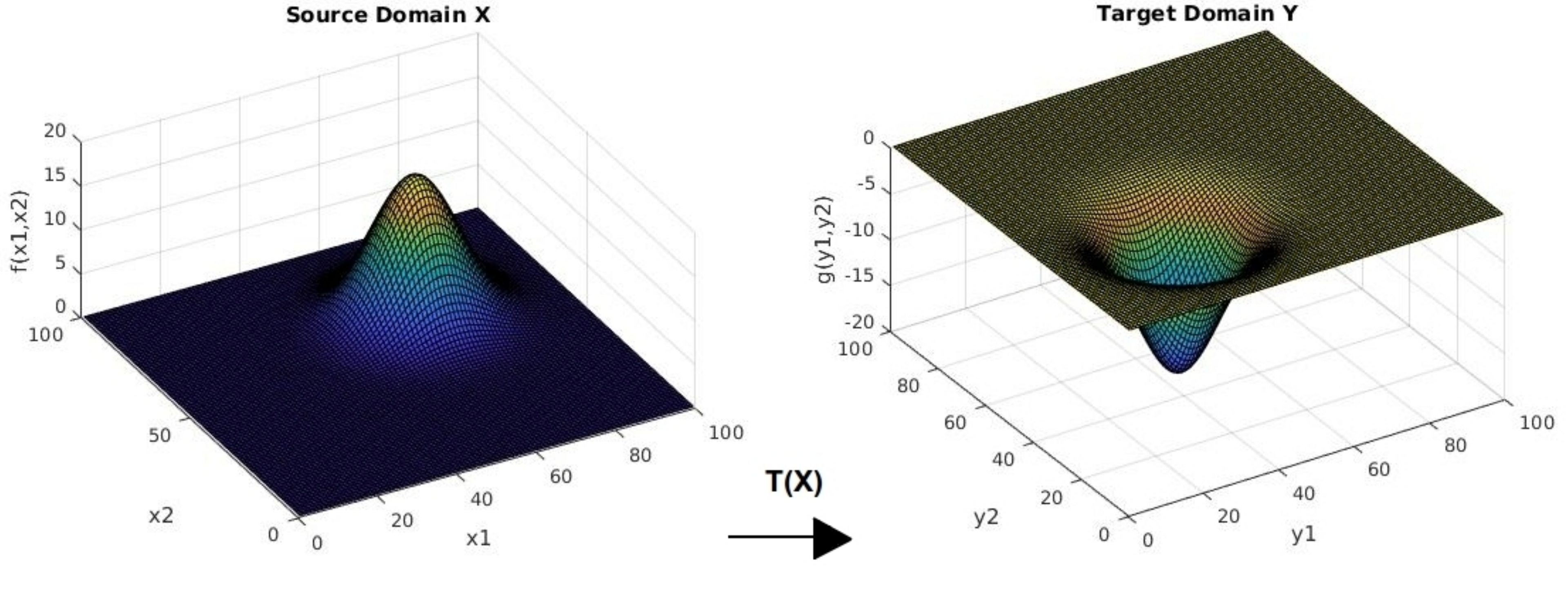}
	\caption{Optimal Transport Motivation}
	\label{dirt}
\end{figure}

The transport problem can be described mathematically as follows. Consider two measure spaces $(X,\mu)$ and $(Y,\nu)$, where $X$ is the source domain and $Y$ is the target domain. The measures $\mu$ and $\nu$ model the pile of dirt and the hole respectively. By normalising such that the volume of the pile of dirt and hole integrate to 1, $\mu$ and $\nu$ can be considered probability measures (more generally Radon measures on a Radon Space). We denote $\mu(A)$ and $\nu(B)$ to represent the mass of measurable subsets $A \subset X$ and $B \subset Y$. Of course we must consider that to transport the mass from some $x \in X$ to some $y \in Y$ requires effort. We prescribe a cost function to measure the effort of transporting mass from $X$ to $Y$. It is natural to consider this cost function as a distance metric and we are describing effort over a distance. So we shall define $c: X \times Y \mapsto [0,+\infty]$, where $c$ is measurable. 

Monge's problem is to find some bijective mapping $T: X \mapsto Y$ that rearranges - or transports - one density to the other. For the mapping $T$ to be valid, we must have that mass is preserved by the mapping, hence

\begin{align}
	\nu(A) = \mu(T^{-1}(A)), \qquad \text{for} \qquad \forall A \subset X.
\end{align}

This is sometimes denoted $\nu = T\#\mu$,  the \textit{push-forward} of $\mu$ by $T$ onto $\nu$. Amongst all such valid maps, we wish to minimise the effort with respect to the cost function and so Monge's problem can be stated as minimising the following:

\begin{align}
I[T] = \int_X c(x,T(x))d\mu(x), \qquad \text{s.t.} \qquad \nu = T\#\mu.
\end{align}

\begin{figure}[H]
	\includegraphics[width=0.5\textwidth, height = 0.15\textheight]{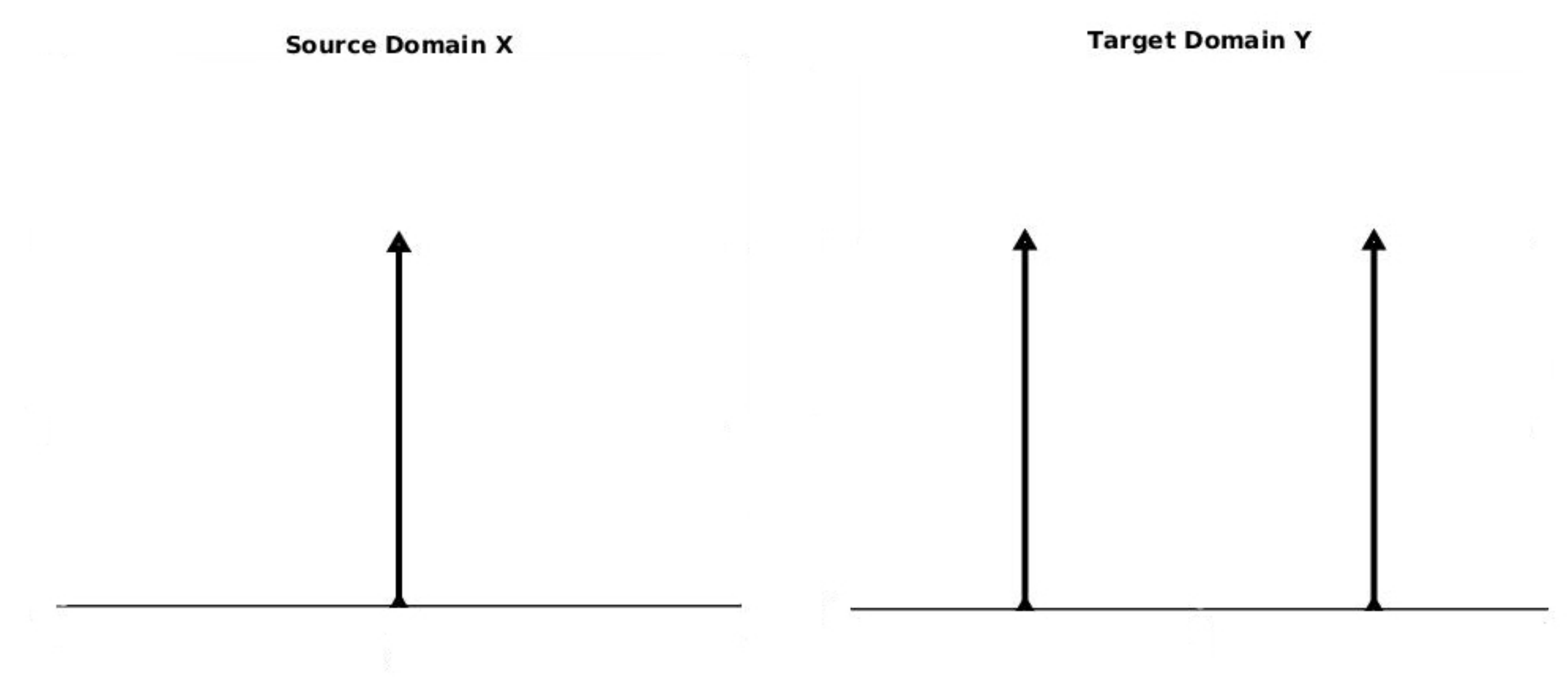}
	\caption{Example where a mapping $T(x)$ cannot exist.}
	\label{Dirac}
\end{figure}

Because of the difficult constraints on the mapping $T$, Monge's problem can be non-trivial to solve and even ill-posed. There may in fact be instances where there does not exist a mapping at all; for example a simple case is to consider when $\mu$ is a Dirac measure $\mu = \delta_0$ but $\nu$ is not. This is illustrated in Figure \ref{Dirac}. 

Trivially, no mapping can exist as mass cannot be split. This also illustrates why Monge's formulation is particularly difficult. Even if valid mappings $T$ do exist, the problem is highly nonlinear and proving the existence of a mapping under different cost functions is potentially very difficult; so we add additional constraints \cite{CV01}. 

We shall assume that $X, Y \subset \mathbb{R}^d$, then for strictly positive,  measurable, Lebesgue densities $f(x)$ and $g(y)$, we have that 

\begin{align}
d\mu(x) = f(x)dx, \qquad \text{and} \qquad d\nu(x) = g(y)dy.
\end{align}

It follows that if $T$ is smooth, then as $T(X) = Y$, by the Jacobian change of variable formula it results in the equation

\begin{align}
\lvert\det(\nabla T(x))\rvert g(T(x)) = f(x),
\end{align}

where $\nabla$ is the gradient operator. This is a fully nonlinear 2nd order Monge-Amp$\grave{\text{e}}$re equation \cite{RC01} and illustrates the highly nonlinear constraints imposed on $T$. 

Monge initially considered the cost function to be proportional to the distance; this actually turned out to be a very difficult case to consider \cite{LA02}. Due to its strict convexity, a simpler case is to consider is when the cost is proportional to the squared distance, i.e. $c(x,y) = \frac{1}{2}\lvert x-y \rvert^2$, when the constant $\frac{1}{2}$ is just for convenience. In this case Monge's problem is to minimise 

\begin{align}
	I[T] = \int_X \frac{1}{2}\lvert \vert x - T(x) \rvert \rvert^2f(x)dx, \quad \text{s.t.} \quad T(X) = Y. 
\end{align} 

When the cost is proportional to the distance squared, Brenier \cite{YB01} has shown that under certain conditions, the optimal mapping is given by the gradient of a convex potential function $\phi: \mathbb{R} \mapsto \mathbb{R}$ such that $\phi = \frac{1}{2}\lvert x \rvert^2 - u$. This is shown in Theorem 1. 

\begin{theorem}
	 There exists a unique mapping of the form $T = x - \nabla u$ that transports $f(x)$ to $g(y)$, this map is also the optimal transport between $f(x)$ to $g(y)$ for the quadratic cost.
\end{theorem}

Substituting this mapping $T(x) = x - \nabla u$ into Equation 3 gives us the equation 

\begin{align} 
\det(I-\nabla^2 u)g(x-\nabla u) = f(x),
\end{align}

subject to boundary conditions satisfying $T(X) = Y$ when mapping a square to itself. For more general domains, the boundary conditions can be significantly more complicated.  As we have assumed $f$ and $g$ are strictly positive probability densities, $\det(I-\nabla^2u)$ must also be strictly positive. By solving the Monge-Amp$\grave{\text{e}}$re equaton (12) to find this optimal mapping, a distance metric naturally arises by considering the cost of transporting the mass under this optimal mapping. The metric is often called the Wasserstein distance, the Monge-Kantorovich distance or the Kantorovich distance; but it is also known by other names, possibly contributing to why it is not so well known in the image processing and machine learning field. For the remainder of this paper we shall call it the Wasserstein distance. The `$p$'th Wasserstein distance is defined by 

\begin{align} 
W_p[T] = \left( \int_X c(x,T(x))f(x)dx\right)^{1/p},
\end{align}

with $p$ depending on the cost function chosen. It can be shown that this metric does indeed satisfy the conditions for being a metric. In our case, for when the cost function is chosen to be proportional to the squared distance, the Wasserstein metric is given by 

\begin{align} 
W_2[T] = \left( \int_X c(x,T(x))f(x)dx\right)^{1/2}.
\end{align}

From this formulation it is possible to compute the distance between probability distributions and indeed images if we consider them as such. In the next section, we present a numerical solution method to the Monge-Amp$\grave{\text{e}}$re equation and hence show a method to evaluate the Wasserstein distance.

\section{A Numerical Solution to the Monge-Amp$\grave{\text{e}}$re Equation}

In this section we shall present a damped Newton numerical scheme designed to give an approximation to the Wasserstein distance between two images. There are two main issues that need to be dealt with to make this approximation sensible. The first main issue is that the PDE formulation assumes smooth, continuous density functions, but images are discrete. This issue is addressed in subsection IV, part B. The second main issue is that it is common for an image representation to have pixels which have an intensity of zero, but in the PDE formulation, we assume strictly positive densities. To resolve this issue, we shall add a small constant to the image before normalisation; this ensures that the image is strictly positive. For our scheme, as we will be experimenting on square images, we shall define our image $\mathcal{I(\zeta)}$ to be an $n \times n$ matrix, with entries $\zeta \in [0+\delta, 1+\delta]$, where $\delta$ is some small constant.

We shall make the assumption that the images to be compared are both the same size. We shall also assume that each image is defined on the unit square, so $X = Y = [0,1]^2$. The unit square is taken to be oriented with positive derivatives in the $x_1$ and $x_2$ directions following the $x_1$ and $x_2$ axis respectively. Despite the assumption for the images to both be square, the scheme is equally valid for images on rectangular domains. 

Then, we solve the two-dimensional Monge-Amp$\grave{\text{e}}$re equation 

\begin{align} 
\det(I-\nabla^2u)g(x - \nabla u) = f(x),
\end{align}

where both $x$ and $u$ are in two dimensions. In the application of images, this is subject to purely homogeneous Neumann conditions at the boundary as we wish to map the edges of each image to the other as well as the corners. The Neumann conditions are given by 

\begin{align*} 
	\dfrac{\partial X}{\partial \mathbf{n}} = 0, \qquad \text{where} \, \mathbf{n} \, \text{is normal to the boundary}
\end{align*} 	

which states no displacement at the boundaries. 

\subsection*{Linearisation}

First we perform a linearisation. To find the linearisation of the Monge-Amp$\grave{\text{e}}$re equation, we make the transformation $u \approx u + \epsilon w$, for small $\epsilon$ and some initial guess $u$. Then (12) becomes 

\begin{align} 
\det(I-\nabla^2(u + \epsilon w))g(x - \nabla (u + \epsilon w)) = f(x).
\end{align}

Denoting partial derivatives by subscripts, e.g. $\partial u / \partial x_1 = u_{x_1}$, for clarity, let 

\begin{align*}\alpha &= 1 - u_{x_1x_1}-u_{x_2x_2} + u_{x_1x_1}u_{x_2x_2} - u_{x_1x_2}^2; \\ \beta &= u_{x_2x_2} - 1; \\ \gamma &= u_{x_1x_1} - 1; \\ \delta &= -2u_{x_1x_2};\end{align*} 

then by expanding (16) while neglecting terms of order $\epsilon^2$ or higher results in

\begin{align*}
\resizebox{.95\hsize}{!}{$\det(I - \nabla^2(u-\epsilon w)) = (\alpha + \beta\epsilon w_{x_1x_1}  + \gamma\epsilon w_{x_2x_2} + \delta\epsilon w_{x_1x_2}).$}
\end{align*}

We take a Taylor series of $g(y)$ around $x-u_x$ such that 

\begin{align*}
	&g(x_1-u_{x_1} - \epsilon w_{x_1},x_2-u_{x_2} - \epsilon w_{x_2}) \approx \\
	 &g(x_1-u_{x_1},x_2-u_{x_2}) - g_{y_1}(x_1-u_{x_1},x_2-u_{x_2})\epsilon w_{x_1} - \\ &g_{y_2}(x_1-u_{x_1},x_2-u_{x_2})\epsilon w_{x_2}.
\end{align*}

For conciseness, let $G = g(x_1-u_{x_1},x_2-u_{x_2})$; that is, $G$ denotes the function $g$ evaluated specifically at the point $(x_1-u_{x_1},x_2-u_{x_2})$. Similarly $G_{y_1}$ and $G_{y_2}$ denote the derivatives of the function $g$ evaluated specifically at the point $(x_1-u_{x_1},x_2-u_{x_2})$. Then our linearised equation is 

\begin{align}&(\alpha + \beta\epsilon w_{x_1x_1}  + \gamma\epsilon w_{x_2x_2} + \delta\epsilon w_{x_1x_2}) \cdot \nonumber \\ 
&(G - G_{y_1}\epsilon w_{x_1} - G_{y_2}\epsilon w_{x_2}) = f(x_1,x_2).
\end{align} 

This is a second order linear partial differential equation in $\epsilon w$. For an initial guess of the solution $u$, we iteratively solve for $\epsilon w$, which is used to update the solution $u_{n+1} = u_n + \epsilon w$. Each iteration can be considered to be a step in a multivariate Newton-Raphson iterative scheme \cite{TY01}. Each iteration of the Newton-Raphson scheme is given by 

\begin{align}
u(\mathbf{x_{n+1}}) \approx u(\mathbf{x_n}) + J_u(\mathbf{x_n}),
\end{align}

where $J_u$ denotes the Jacobian matrix defined over the function $u$. In our numerical scheme we can see that $\epsilon w$ corresponds to the Jacobian. In practice, we iteratively solve until a certain tolerance level is met based on the residual. We also introduce a damping parameter $\gamma$, for $\gamma \in (0,1]$. In some cases, for example if the initial guess for $u$ is poor, the Newton-Raphson method may fail to converge. One possible fail-safe is to use a damping parameter to decrease the contribution from each iteration; the drawback is this will increase the number of iterations.

\subsection*{Discretisation}

To solve $(17)$ numerically, we consider standard central differences on a uniformly discretised grid. In the case of images, we consider an image to be a discretisation of a surface on $[0,1]^2$. Hence, the intensities at the pixels of the image coincide with the discretisation of the grid. This makes sense as an approximation because the image is well defined at these points. Because of this choice, an issue does arise in the target image. When evaluating the function $G$ and its derivatives in $(17)$, the evaluation point $(x_1-u_{x_1},x_2-u_{x_2})$ will generally be unknown as this point will very likely not lie on a known pixel point. To solve this issue, we perform an interpolation on the pixels to construct a continuous function from the discrete representation of the image. For our choice of interpolation, we have used a spline approximation; although other choices can be justified depending on the type of image one deals with. 

The standard central differences are set out as follows. For a general function $\psi$, we denote its position on the grid by $\psi_{i,j}$, for $i = 1, 2, \, ... \, , m$ and $ j = 1, 2, \, ... \,, n$, where $m$ and $n$ are the number of grid points in the $x_1$ and $x_2$ directions respectively. The following central differences are used with $h$ and $k$ the step sizes in the $x_1$ and $x_2$ directions respectively:

\begin{align}
	\psi_{x_1} &= \dfrac{\psi_{i+1,j} - \psi_{i-1,j}}{2h}; \nonumber \\
	\psi_{x_2} &= \dfrac{\psi_{i,j+1} - \psi_{i,j-1}}{2k}; \nonumber  \\
	\psi_{x_1x_1} &= \dfrac{\psi_{i-1,j} - 2\psi_{i,j} + \psi_{i+1,j}}{h^2};  \\
	\psi_{x_2x_2} &= \dfrac{\psi_{i,j-1} - 2\psi_{i,j} + \psi_{i,j+1}}{k^2};  \nonumber \\
	\psi_{x_1x_2} &= \dfrac{\psi_{i+1,j+1} - \psi_{i+1,j-1} - \psi_{i-1,j+1} + \psi_{i-1,j-1}}{4hk}.  \nonumber 
\end{align}

By multiplying out (17) and again neglecting terms of order $\epsilon^2$ or higher results in 

\begin{align}
	&G(\beta\epsilon w_{x_1x_1}  + \gamma\epsilon w_{x_2x_2} +\delta\epsilon w_{x_1x_2}) \nonumber \\ 
	&-\alpha(G_{y_1}\epsilon w_{x_1} + G_{y_2} \epsilon w_{x_2}) = f(x_1,x_2) - \alpha G.
\end{align}

By discretising $\epsilon w$ with the central finite differences outlined in (19), the equation for an interior point in the grid is given by

\begin{align}
&-2G\left(\dfrac{\beta}{h^2}+ \dfrac{\gamma}{k^2}\right) \epsilon w_{i,j} + \left(\dfrac{\beta G}{h^2} + \dfrac{\alpha G_{y_1}}{2h}\right) \epsilon w_{i,j-1} \nonumber \\ & \qquad + \left(\dfrac{\beta G}{h^2} - \dfrac{\alpha G_{y_1}}{2h}\right) \epsilon w_{i,j+1} \, \nonumber \\ & \qquad +  \left(\dfrac{\gamma G}{k^2} + \dfrac{\alpha G_{y_2}}{2k}\right)\epsilon w_{i-1,j} \\ & \qquad + \left(\dfrac{\gamma G}{k^2} -  \dfrac{\alpha G_{y_2}}{2k}\right) \epsilon w_{i+1,j} \, + \nonumber \\ &\dfrac{G\delta }{4hk}(\epsilon w_{i+1,j+1} -  \epsilon w_{i+1,j-1} - \epsilon w_{i-1,j+1} + \epsilon w_{i-1,j-1}) \nonumber \\ & \qquad = f_{i,j} - \alpha G, \nonumber
\end{align}

where each of the coefficients are known values from our initial guess for $u$.

For points on the boundary, ghost points (see for example \cite{RW01}) can be considered and eliminated using the central difference formulae. For example, for a point $p$ on the left-hand boundary $(1,p)$, the following two equations can be used: 

\begin{align*}
	\psi_{x_1x_1}(1,p) &= \dfrac{\psi_{0,p} - 2\psi_{1,p} + \psi_{2,p}}{h^2}; \\
	\psi_{x_1}(1,2) &= \dfrac{\psi_{2,p} - \psi_{0,p}}{2h}.
\end{align*}

Imposing the Neumann condition $\partial X / \partial \mathbf{n} = 0$, we have that $
(\psi_{2,p} - \psi_{0,p})/2h = 0 \implies \psi_{0,p} = \psi_{2,p}$. Hence we can eliminate the ghost point $\psi_{0,p}$ from the second derivative equation. Analogously, the same treatment can be used in the $x_2$ direction and trivially, the mixed derivatives are zero on the boundary. This leads to system of linear equations, which in matrix form are represented as

\begin{align} 
A\mathbf{\epsilon w} = b,
\end{align}

where $A$ is an $n \times n$ coefficient matrix resulting from the left-hand side of (21), and $b$ is the right-hand side of (21). From imposing purely homogeneous Neumann conditions, the coefficient matrix has a rank of $n-1$. This means the linear system is underdetermined; there is one more variable than the number of equations. One possible solution is to set one of the variables to some value to eliminate a row (or equivalently column) of $A$. This is rather arbitrary without an idea of what the solution $u$ looks like. Instead, we impose an extra condition such that solution is orthogonal to the vectors in the kernel, where the kernel, also known as the null space, is the set of the solutions that solves $A\epsilon \mathbf{w} = 0$. This corresponds to the bordered system

\begin{align}
\begin{bmatrix}
A \, \, e \\ e' \, \, 0
\end{bmatrix}
\begin{bmatrix}
\epsilon w \\ \lambda
\end{bmatrix}
= 
\begin{bmatrix}
b \\ 0
\end{bmatrix}
,
\end{align}

where $e$ is the vector of ones. In this case, by the rank-nullity theorem \cite{CM01}, we know that kernel of this system is just a single vector and so the solution is orthogonal to just this single vector. This system can be solved and the solution $u$ can be updated accordingly. In practice, we make a sensible initial guess of $u = 0$ for the first iteration which corresponds to the identity mapping and choose the tolerance test to be the absolute difference between the maximum and the minimum of the residual. 

\begin{figure}[H]
	\begin{center}
		\includegraphics[height=0.65\textheight, width=0.5\textwidth]{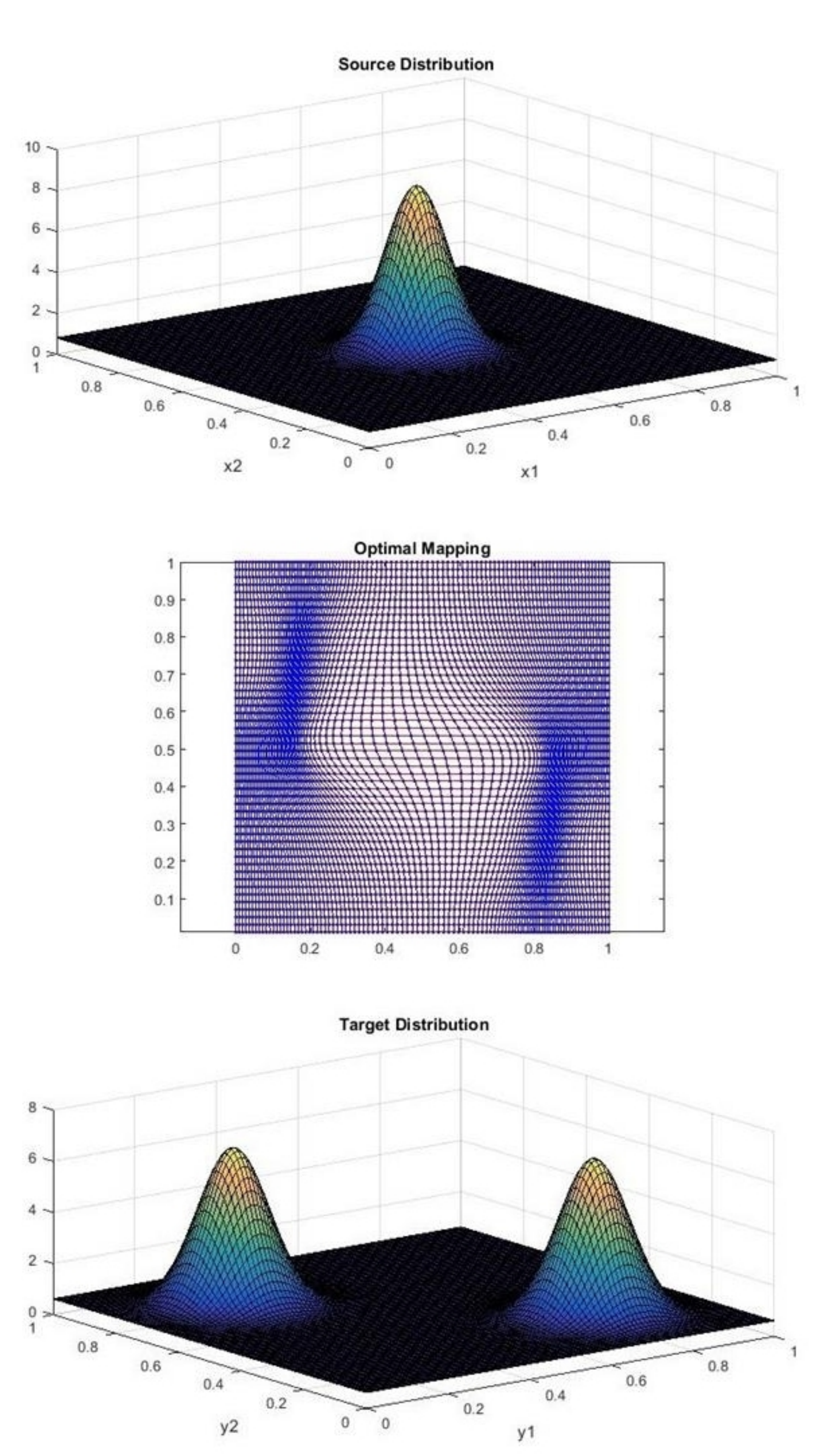}
		\caption{An example of an optimal map on a mesh grid.}
		\label{optimalmapping}
	\end{center}
\end{figure}

Finally, after the solution $u$ is found, subject to some threshold of tolerance, the optimal mapping $T(x)$ can be evaluated from Theorem 1. The Wasserstein distance can also then be computed as $T(x)$ is known; so we perform a numerical integration of 

\begin{align} W = \left( \int_X \frac{1}{2}\lvert \lvert x - T(x) \rvert \rvert f(x) dx \right)^{1/2},
\end{align}

using an appropriate method. In our algorithm we have chosen to use the trapezium rule. As an example of the mapping, we can consider an image which represents a Gaussian centred in the unit square and compare that to an image of two Gaussians centred in opposing corners of the unit square. Figure \ref{optimalmapping} illustrates the effect of the mapping on a uniform mesh grid.

\subsection*{Computational Complexity}

\begin{figure}[H]
	\begin{center}
		\includegraphics[width=0.55\textwidth, height=0.35\textheight]{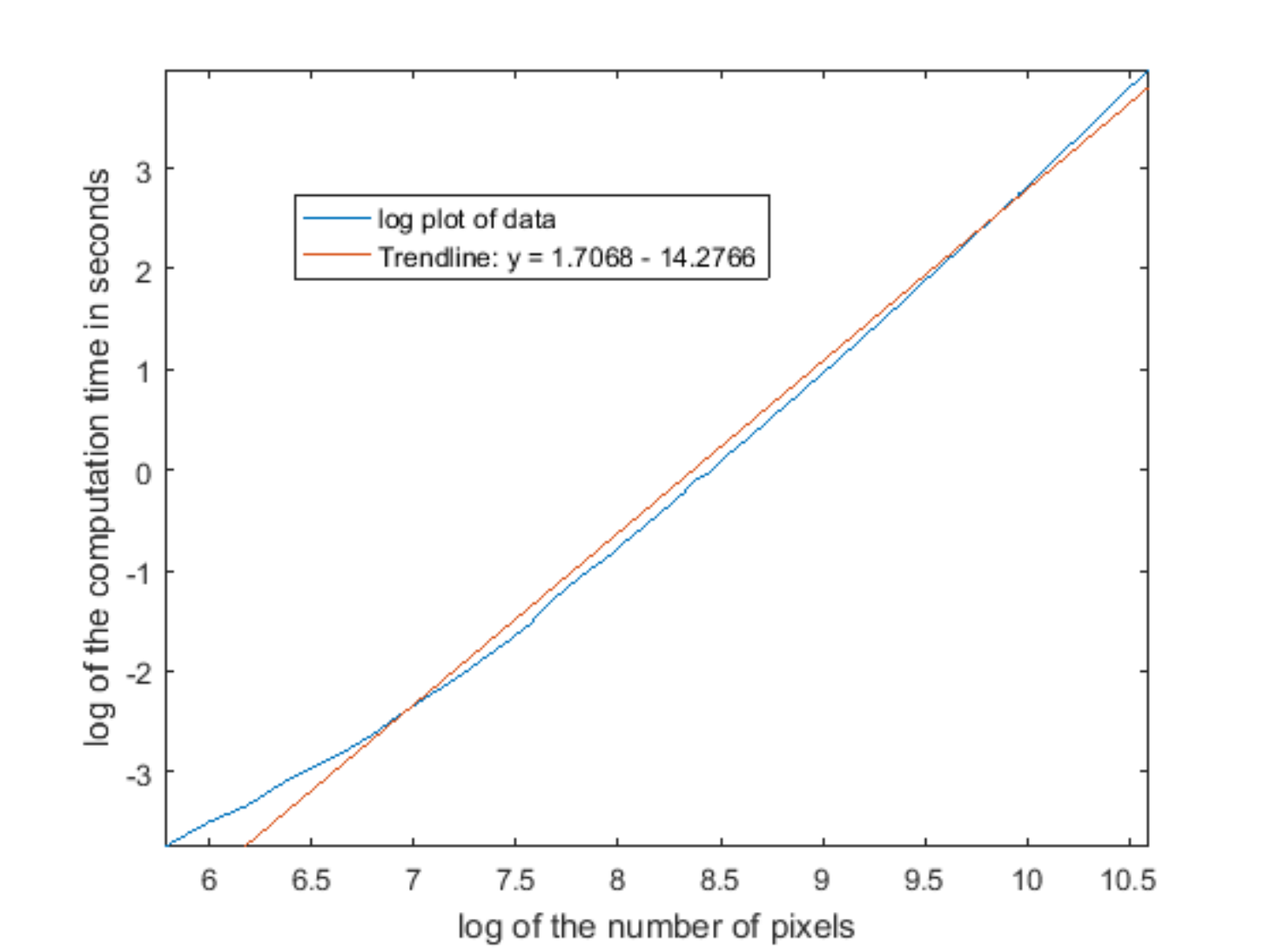}
		\caption{A log-log plot of computation time (secs) against the total pixel number.}
		\label{complexity}
	\end{center}
\end{figure}

For this algorithm, there are three main areas of computational complexity: setting up the sparse matrix; solving the linear system; and the number of iterations needed to reach the tolerance threshold. The main source of complexity is solving the linear system which is potentially large. Solving discretised elliptical PDEs is a well understood problem; due to the structure of the coefficient matrix, solvers are able to take exploit the sparsity. Typical solvers will be able to solve such a system with a complexity of approximately $O(N\log(N))$ \cite{AG01} and in some situations, faster. With the added complexity of the amount of iterations and setting up the matrix, we would expect our method to have a complexity somewhere between $O(N)$ and $O(N^2)$. Figure \ref{complexity} illustrates an experiment where we map a Gaussian density centred in one corner of an image to a Gaussian density centred in the opposite corner of the target image; which is a complex mapping. Figure \ref{complexity} is a log-log plot of the computational time in seconds against the total number of pixels. 

Indeed, in this case, the complexity falls between $O(N)$ and $O(N^2)$; roughly $O(N^{1.7})$, although this may differ for more images with a more complicated geometry where the nonlinear solver has to compute more iterations to reach the threshold; and conversely simpler images where the nonlinear solver reaches the threshold in fewer iterations. The algorithm can also be made more efficient by introducing a variable damping parameter to reach the desired threshold in fewer steps.

\section{$k$-Nearest Neighbours Classification}

One of the simplest machine learning algorithm is the $k$-nearest neighbours algorithm. The $k$-NN algorithm is an example of non-parametric instance-based learning \cite{NA01}; given a set of labelled training data and a set of labelled test data, each instance of test data is compared to every instance of training data subject to some similarity measure. The $k$-nearest neighbours are chosen and the data is classified via a majority voting system. 

More formally, we consider general data points to be in a metric space $(X,d)$, equipped with some distance $d$, where each data point has an associated class. We choose some $k \in \mathbb{N}^+$ to define the number of nearest neighbours that influence the classification. In the simplest case $k = 1$, the classification rule is simply the training vector closest to the test vector under the distance $d$. 

\subsection{Distances}

In our experiment we shall use the $k$-nearest neighbours algorithm to compare the Wasserstein distance $(8)$ with two widely used distance measures. The first we have chosen is the well-known Euclidean distance; given two n-dimensional vectors to compare $\mathbf{x_n}$ and $\mathbf{y_n}$, the Euclidean distance is defined as

\begin{align*}
d(\mathbf{x}, \mathbf{y}) = \sqrt{(x_1 - y_1)^2 + (x_2 - y_2)^2 + \cdot \cdot + (x_n - y_n)^2}.
\end{align*}

As noted in Section I, we may expect the Euclidean distance to perform badly in some cases as a standalone metric. But in the case of $k$-nearest neighbours it can be effective with a large training set given that you may expect at least one training example to be similar to a test example. The second measure we shall use is Pearson's linear correlation coefficient to between the matrices representing the images \cite{KP01}. Pearson's coefficient gives a measure of the linear dependence between two variables and returns a scalar $p \in [-1,1]$. The $p$-value is a scale where $p=-1$ is a perfect negative linear trend, which corresponds to the inverse of the image, and $p=1$ is a perfect positive linear trend. A value of $p=0$ indicates no correlation. For two images $x$ and $y$, Pearson's correlation coefficient is defined as 

\begin{align}
p = \dfrac{\sum_i (x_i - \overline{x})(y_i - \overline{y})}{\sqrt{\sum_i (x_i - \overline{x})^2} \sqrt{\sum_i (x_i - \overline{x})^2}},
\end{align}

where $x_i$ is the $i$th pixel of $x$ and $\overline{x}$ denotes the image mean; that is, the mean of all of the pixels in the image. The significant difference between the Euclidean distance and Pearson's coefficient is that the latter is invariant to both scale and location transformations, making it a potentially more robust measure. In our experiment we shall use the absolute value of of the Pearson coefficient as a measure of similarity between the two images as we are interested in a relationship between the images, whether positive or negative. The last distance we shall compare against is the Tangent distance space, which produces state-of-the-art results on optical character recognition datasets with the $k$-NN framework. Simard et al. \cite{PS01} developed a distance measure that is invariant to small transformations of images, something that other distances such as the Euclidean distance are very sensitive to. Simard et al. approximated the small transformations by creating a tangent space to the image by adding the image $\mathcal{I}$ to a linear combination of the tangent vectors $t_l(x)$, for $l = 1,.\, . \, , L$, where $L$ is the number of different transformation parameters. The tangent vectors are the partial derivatives of the the transformation with respect to the parameter and span the tangent space. So the tangent space is defined as

\begin{align} 
M_x = \{ \mathcal{I} + \sum\limits_{l = 1}^L t_l(x)\cdot \alpha : \alpha \in \mathbb{R}^L \},
\end{align}

where $\alpha$ is the transformation parameter, for example the scale or translation. The one-sided Tangent space distance is they defined as 

\begin{align}
TSD\left(\mathcal{I}_1, \mathcal{I}_2\right) = \min\limits_{\alpha} \{ \lVert \mathcal{I}_1 + \sum\limits_{l = 1}^L t_l(x)\cdot \alpha - \mathcal{I}_2 \rVert \}.
\end{align}

In our experiment we have used Keyser's excellent implementation of the Tangent space distance \cite{DK01}.

\section{Nearest Neighbour Classification on the MNIST dataset}

\begin{figure}[H]
	\begin{center}
		\includegraphics[width=0.45\textwidth]{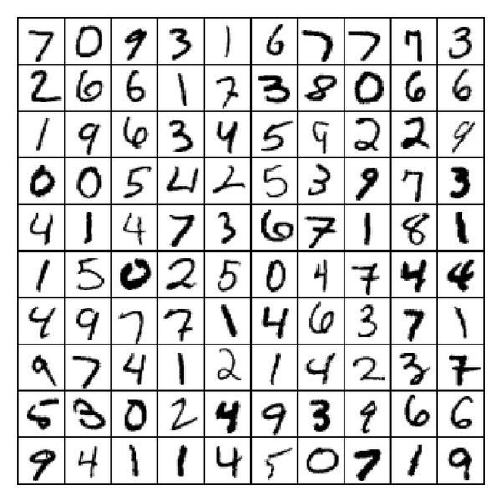}
		\caption{Examples of MNIST Digits}
		\label{mnist1}
	\end{center}
\end{figure}

The MNIST digit dataset is one of the most well-known datasets in digit recognition applications. It was constructed by LeCun et al. \cite{YL01} from the NIST dataset. According to LeCun et al. the original binary images were size normalised to fit into a 20 $\times$ 20 pixel box while preserving their aspect ratio. The MNIST dataset are a grey level representation of the images, resulting from the anti-aliasing algorithm used in their normalisation. The images are centred into a $28 \times 28$ image by computing the centre of mass of the pixels and translating the images such that this point coincides with the centre of the $28 \times 28$ image. The dataset contains a training set with 60,000 images and a test set of 10,000 images. This is a convenient dataset to test our algorithm on as it requires minimal preprocessing and the images are centred so we can focus on the distance metric's discriminatory power over translation and scaling invariance. Figure \ref{mnist1} examples some of the digits. Although this is a nice dataset in that it requires minimal preprocessing, it is worth noting that this is not the only choice one can make in the preprocessing steps. For example, if the images are not centred by their mass and just drawn within some bounding box, the results of a nearest neighbour classification may wildly differ depending on the properties of the distance metric chosen.

\begin{figure}[H]
	\begin{center}
		\includegraphics[width=0.45\textwidth]{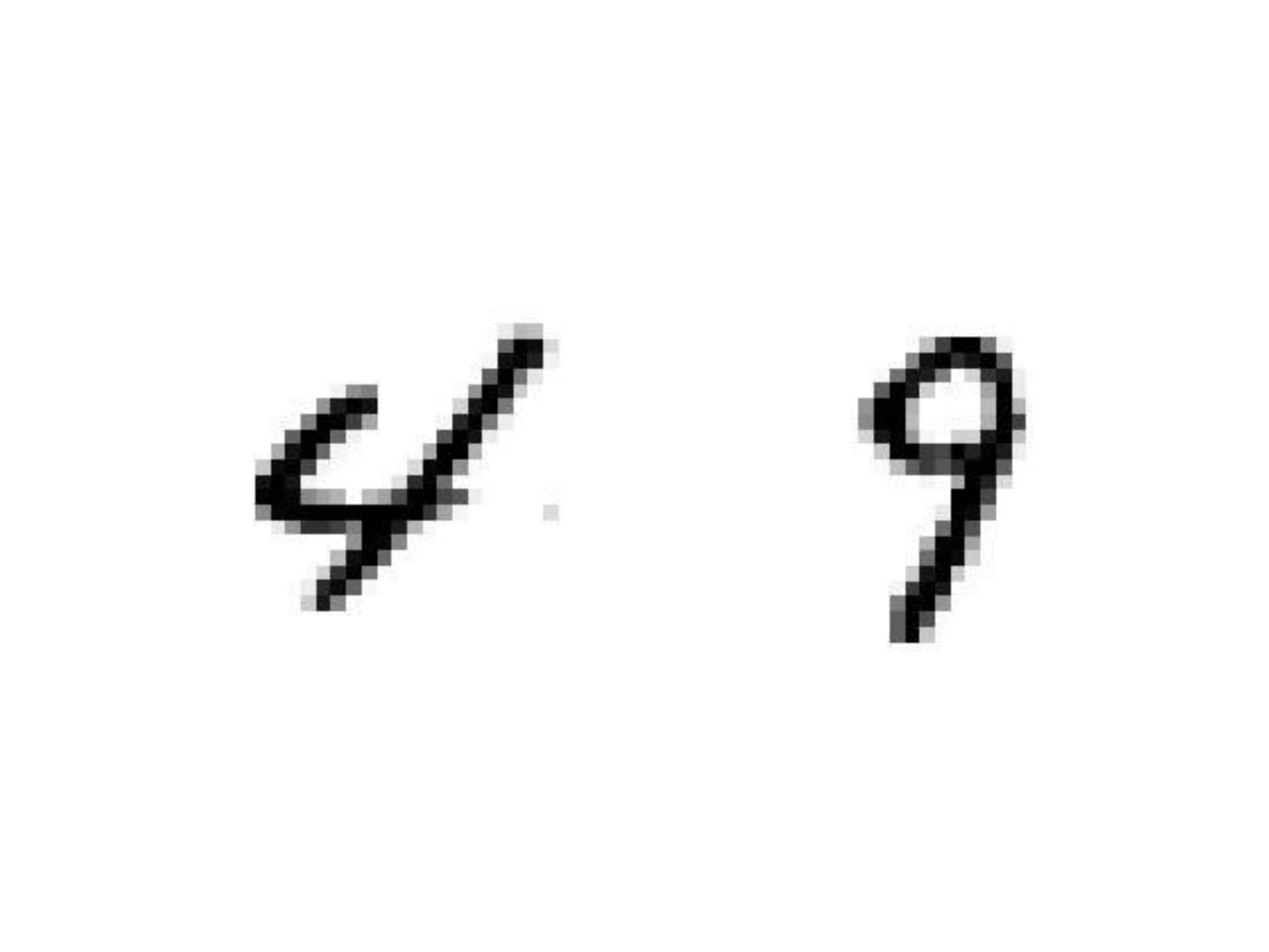}
		\caption{Example of similar digits}
		\label{mnist}
	\end{center}
\end{figure}

This dataset is a strong test of how well a distance metric can perform given the natural deviations in how people write digits. In particular, some of the digits can be very similar in pixel distribution. Consider for example the digits $9$ and $4$ seen in Figure \ref{mnist}; although we can clearly see the difference in the digits, they may be hard to differentiate using a distance comparison algorithm. This is an example of why a distance metric must be able to discriminate well. This will be less important in very large training sets where it is likely at least one piece of data in the training set will be similar to the test data. But for smaller training sets, the discriminatory power of the distance can be more powerful. For example, performing a $k$-NN classification equipped with the Euclidean distance on the whole MNIST dataset returns an error rate of 3.12$\,\%$.

In our experiment, for both efficiency and due to a lack of computational power, we decided to take subset of the MNIST dataset to perform our $k$-NN test. We first randomised the original dataset and then created a training set of 10,000 images of which contained 1,000 different examples of each of the digits. This was further partitioned into 20 disjoint training sets containing 50 samples of each digit. A test set was randomly constructed from the MNIST test set which contained 200 images, of which there were 20 of each digit. Both the training set and the test sets were normalised such that each image numerically integrated to 1. 

We were interested to see how each metric performed as the training set increased in size, starting from a very small training set (see for example \cite{AS01}). To that end we split our experiment into 25 different tests; for the initial test we performed the $k$-NN algorithm with a training set of just 1 of each digit. We repeated this test on each of the three distances to be tested, each time increasing the number of each digit in the training set by 1. E.g. for the final test, we had a training set containing 25 of the digit `one', 25 of the digit `two' and so on, for a training set of size 250. This experiment was then tested on each of the 20 different constructed training sets to take into account the variability of the data sets. The reason we have not tested on the whole data set of 50 is that we had limited computing power, but we have included the result of the whole training set being tested. 

In this experiment, we shall impose the same conditions on each of the distance measures. In each case in the $k$-NN algorithm we define $k$ to be $1$, so that the classification is based on just the nearest neighbour. In the case of the Wasserstein distance, to ensure that our numerical solution method works for every image with the same parameters, we add a constant of $1$ to each image before normalisation to ensure every image is non-zero. In this case, we set the damping parameter in our update step $u \approx u + \phi\epsilon w$ to be just $1$. 

\begin{figure}[H]
	\begin{center}
		\includegraphics[width=0.50\textwidth, height = 0.35\textheight]{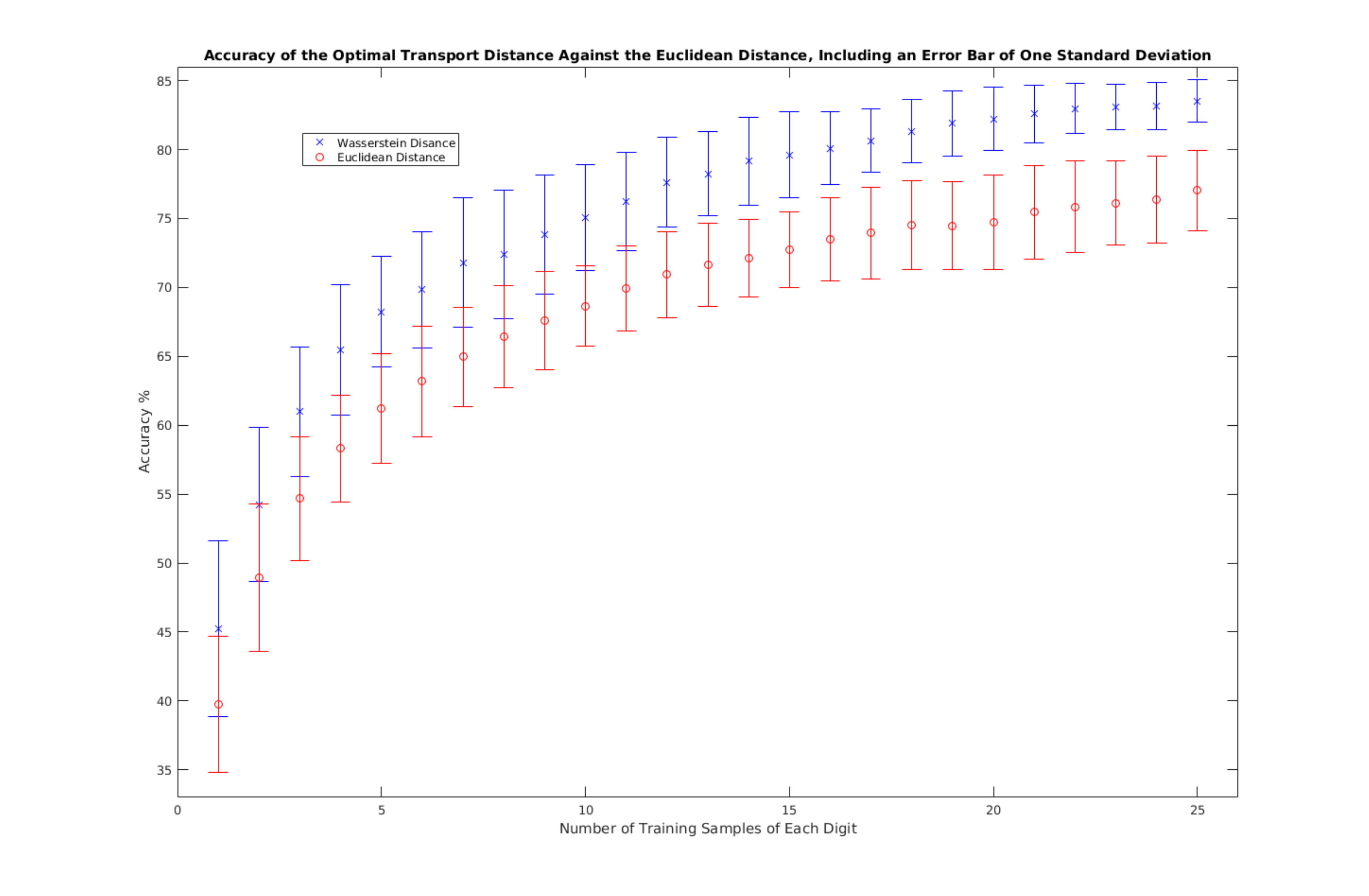}
		\caption{- An average of the results for the Wasserstein and Eulcidean distance over all 20 training sets; including an error bar of one standard deviation is used.}
		\label{Euclid_results}
	\end{center}
\end{figure}

\section{Results}

\begin{figure}[H]
	\begin{center}
		\includegraphics[width=0.5\textwidth, height = 0.32\textheight]{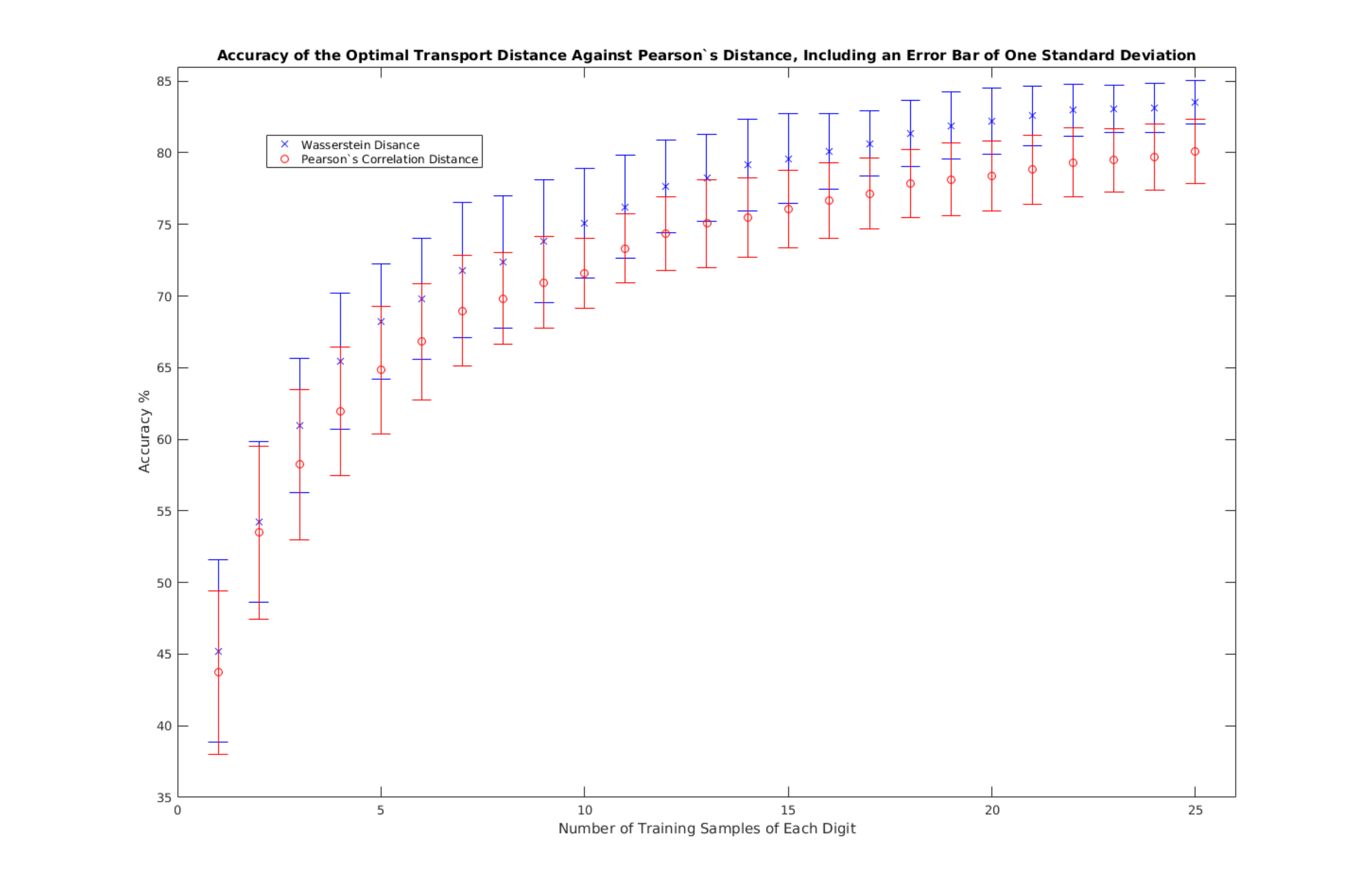}
		\caption{- An average of the results for the Wasserstein and Pearson's Correlation distance over all 20 training sets; including an error bar of one standard deviation is used.}
		\label{Pearson_results}
	\end{center}
\end{figure}

In this section we shall present our main results. The main result we captured was the accuracy of each tested distance averaged over all 20 of the disjoint training sets. This main result is very promising: the Wasserstein distance outperforms each of the Euclidean distance, the distance defined from Pearson's correlation coefficient and the Tangent space distance. For a training set of just 25 of each digit, the Wasserstein distance already achieves an accuracy of $83.525$$\,\%$, which is a very encouraging result considering the size of the training set is less than $1$$\,\%$ of the whole MNIST training set. In comparison, the Tangent space distance achieves an accuracy of $82.2$$\,\%$; Pearson's correlation distance performs at $80.1$$\,\%$ accuracy; and the Euclidean distance achieves a $77.375$$\,\%$ accuracy. The results over each of the 25 different training set sizes are presented in Figures \ref{Euclid_results}, \ref{Pearson_results}, and \ref{Tangent_results} which each include an error bar of one standard deviation.

\begin{figure}[H]
	\begin{center}
		\includegraphics[width=0.50\textwidth, height = 0.32\textheight]{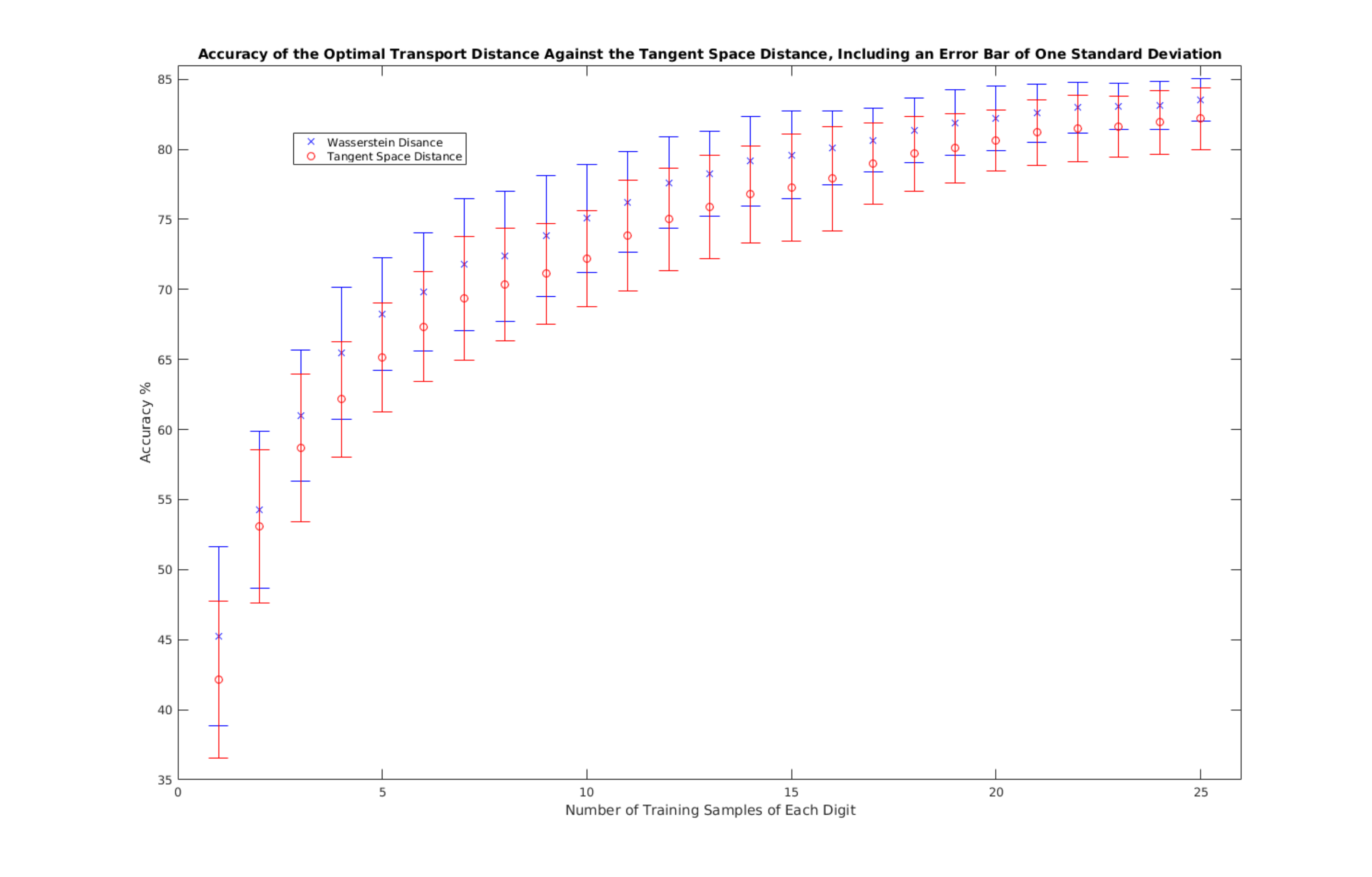}
		\caption{- An average of the results for the Wasserstein and the Tangent Space distance over all 20 training sets; including an error bar of one standard deviation is used.}
		\label{Tangent_results}
	\end{center}
\end{figure}

We also include a sample of the results in Table 1 to complement the graphics illustrating the results. In Table 1 we also include the average accuracy of all 20 tests tested on the whole training set of 50 of each digit to show how well the distances are performing on a larger dataset. From the results Figures, we can see that the accuracy graphs of the distances each exhibit asymptotic behaviour; as the number of training examples increases towards the full MNIST training set, we expect each distance to converge to some value below 100\%. We can conjecture that for a very large training set, for example the original MNIST training set, the Wasserstein distance will have a better performance to each of the three tested distances. With additional computational power we would like to test this hypothesis on the whole MNIST dataset and see how this compares to other high-performing distance metrics. It is worth noting that the choice of distance metric is dependent on the type of application and data you are looking at. The optimal transport distance is a very natural way to compare images, so in some applications, it is certainly worth consideration. 

\begin{figure}[H]
	\begin{center}
		\includegraphics[width=0.45\textwidth, height = 0.12\textheight]{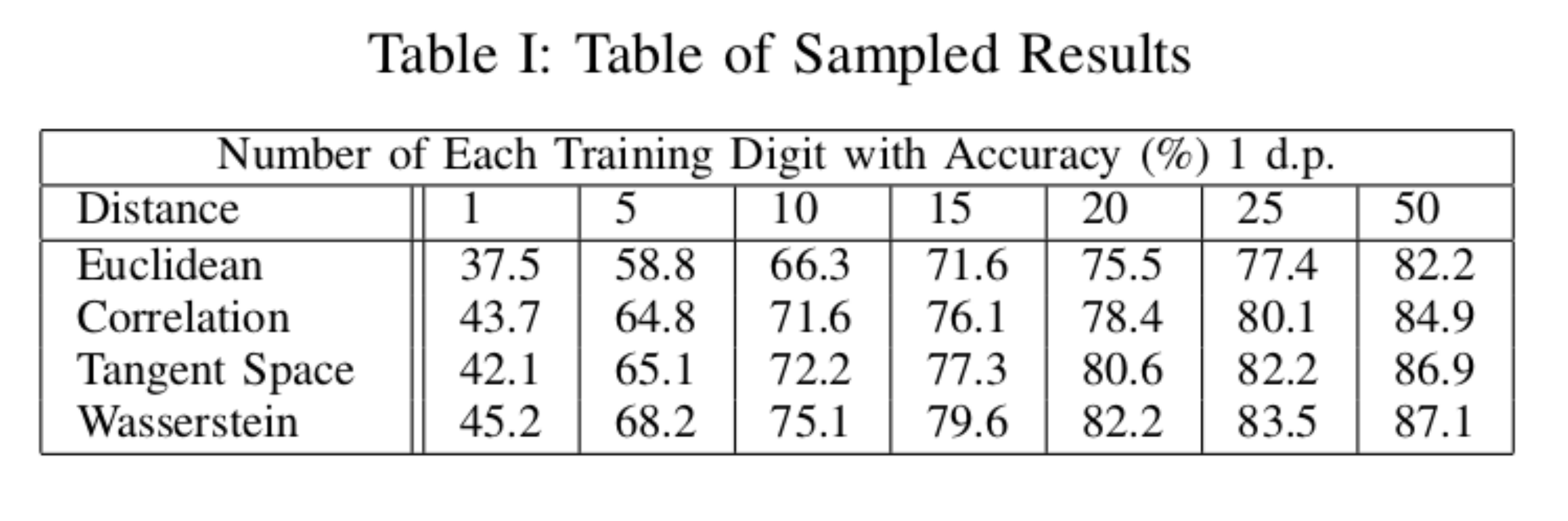}
		\label{Tangent_results}
	\end{center}
\end{figure}

As the results indicate that the Wasserstein distance can perform well with just a few training examples, one possible approach would be represent each class in the training set by just one representation and see how the different metrics perform in this case. These results suggest that this kind of approach could be fruitful. We can conclude that our implementation of the PDE formulation of Monge's optimal transport distance has shown some very encouraging results when experimenting on the MNIST dataset. This experiment illustrates the real potential of the Wasserstein distance in image processing and wider machine learning applications. 

\section{Summary and Future Work}

This paper presents an implementation of Monge's optimal transport problem with its associated Wasserstein distance. The Wasserstein distance is used as the distance metric within the $k$-nearest neighbours algorithm and we have presented some comparative results on a subset of the MNIST dataset against two widely used distance measures. The results are very promising with the Wasserstein distance outperforming both the Euclidean distance and Pearson's coefficient. This demonstrates that in some situations it may indeed be more beneficial to use the Wasserstein metric over other traditional methods. In particular, in the application of machine learning, we expect that when the training set is limited, the Wasserstein distance could potentially perform better than other metrics due to its discriminatory versatility. 

The drawback to using this metric is both the non-trivial implementation and potentially the computational cost. For larger images, the algorithm could be prohibitive dependant on the user's needs; so the user would have to decide on whether the metric's benefits outweigh the computation time. For future work, we would like to improve the efficiency of the algorithm by using a more sophisticated and flexible nonlinear solver. We would also like to investigate the accuracy of the algorithm when representing the image by different continuous representations; for example using linear or cubic interpolation as opposed to spline approximations. We would also like to implement the discrete Kantorovich distance, as described in Section I, to compare accuracy to the PDE implementation. The benefit of the discrete formulation is that as it reduces to a linear programming problem, the formulation is very neat and the subject of linear programming is very well understood. This approach also lends itself naturally to images as it is a discrete metric; but the drawback is again computation time.

\bibliography{/home/mike/Dropbox/Latex/arXiv-submissions/mybib_v2}

\bibliographystyle{ieeetr}

\end{multicols}
\end{document}